\documentclass[runningheads]{llncs}
\usepackage[T1]{fontenc}
\usepackage{graphicx,verbatim}
\usepackage{graphicx}
\usepackage{cite} 
\usepackage{times}
\usepackage{epsfig}
\usepackage{amsmath}
\usepackage{amssymb}
\usepackage{mwe}
\usepackage{acro}
\usepackage{amssymb}
\usepackage{xcolor,colortbl}
\usepackage{tabularx}
\usepackage{relsize}
\usepackage{pifont}
\usepackage{booktabs} 
\usepackage{multirow}
\usepackage{multicol}
\usepackage{adjustbox}
\usepackage{float}
\usepackage[colorlinks,
            linkcolor=red,  
            anchorcolor=blue, 
            citecolor=green,   
            ]{hyperref}
\usepackage[flushleft]{threeparttable}

\begin{document}
\title{MORI-Seg: Learning Morphological Geometry for Instance Segmentation without Instance Annotations}
\titlerunning{MORI-Seg}

%

\author{
Leiyue Zhao\inst{1} \and
Tianyu Shi\inst{2}	\and
Daniel Reisenb\"uchler\inst{3}	\and
Xinzi He\inst{4}	\and
Junchao	Zhu\inst{5} \and
Tianyuan Yao\inst{5} \and
Yuechen	Yang\inst{5} \and
Yanfan	Zhu\inst{5} \and
Junlin	Guo\inst{5} \and
Gelei	Xu\inst{6} \and
Haichun	Yang\inst{5} \and
Yuankai Huo\inst{5,7} \and
Mert R. Sabuncu\inst{4,8,9} \and
Yihe Yang\inst{8,9} \and
Ruining Deng\inst{5,9}\thanks{Corresponding author.\email{rud4004@med.cornell.edu}}
}


\authorrunning{L. Zhao et al.}

\institute{
Southern University of Science and Technology, Shenzhen, Guangdong, 518055, China \and
Sichuan University, Chengdu, Sichuan, 610064, China\and 
University of Regensburg, Regensburg, Bavaria, 93053, Germany \and
Cornell University, Ithaca, NY, 14853, USA \and
Vanderbilt University, Nashville, TN, 37235, USA \and
University of Notre Dame, Notre Dame, IN, 46556, USA \and
Vanderbilt University Medical Center, Nashville, TN 37232, USA \and
Cornell Tech, New York, NY 10044, USA \and
Weill Medical College of Cornell University, New York, NY 10065, USA 
}

\maketitle              
\begin{abstract}
Instance-level quantification of kidney functional units is essential for morphometric analysis, yet most publicly available pathology datasets provide only semantic segmentation annotations, where adjacent structures of the same class are merged into single regions. This prevents reliable instance-level analysis and limits downstream quantitative studies. Existing heuristic post-processing methods often yield suboptimal instance separation, particularly in crowded and adherent regions, while deep learning-based instance segmentation approaches typically require intensive instance-level annotations that are costly and labor-intensive to obtain. We propose \textbf{MORI-Seg}, a deep learning framework that enables instance segmentation without requiring instance-level annotations. Instead of heuristic splitting or instance supervision, MORI-Seg learns morphology-aware geometric representations directly from semantic masks by jointly modeling object-centric distance fields and boundary-band representations to encode interior structure and contact interfaces. A class-conditioned feature disentanglement module further promotes intra-instance coherence and inter-instance separation. Under semantic-only supervision, MORI-Seg decomposes connected semantic regions into distinct instance masks in an end-to-end manner. Experiments demonstrate improved instance separation accuracy and more reliable morphometric quantification compared with classical post-processing pipelines and representative semantic-to-instance learning approaches. The official implementation is publicly available at \url{https://github.com/ddrrnn123/MORI-Seg}.


\end{abstract}
\section{Introduction}

\begin{figure*}[t]
        \centering
        \includegraphics[width=0.9\linewidth]{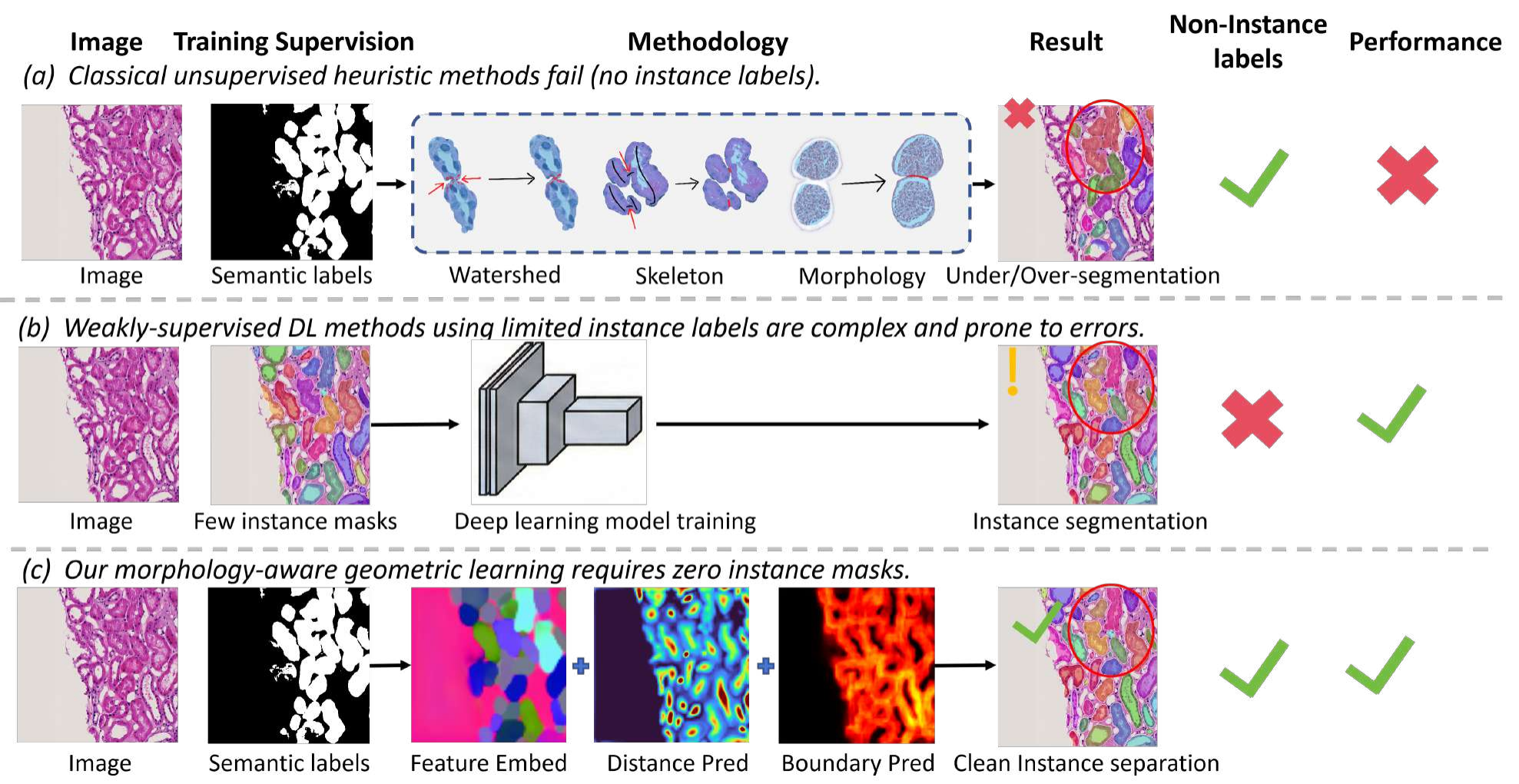}
        \caption{\textbf{From semantic masks to instance segmentation.} 
(a) Heuristic post-processing requires no instance annotations but often yields under- or over-segmentation in crowded kidney tissues. 
(b) Learning-based methods achieve accurate separation but rely on costly instance masks. 
(c) \textbf{MORI-Seg} bridges this supervision gap by learning morphology-aware geometric priors from semantic masks alone, enabling robust instance-mask-free instance segmentation.}
        \label{fig:Problem}
\end{figure*}

Instance-level quantification of kidney functional units is essential for morphometric analysis and its association with clinical outcomes~\cite{gupta2019emergence}. Measuring the size, shape, density, and spatial organization of biological structures requires reliable quantitative segmentation and structural characterization~\cite{yang2025pyspatial,chen2024spatial, zhou2024innovations}. As digital pathology datasets continue to expand, scalable and reproducible instance-level analysis becomes increasingly important~\cite{xiong2025advances,bouteldja2021deep,shi2026m}.

However, most publicly available kidney pathology datasets provide only semantic segmentation annotations~\cite{jayapandian2021development,deng2025kpis}. In semantic masks, multiple adjacent structures of the same class are merged into single connected regions, preventing direct instance-level quantification~\cite{wang2025glo,altini2020semantic,deng2025casc,deng2024hats,deng2025segment}. This creates a fundamental supervision gap: instance segmentation is required for quantitative analysis, yet instance-level annotations are often unavailable due to the substantial manual effort required for delineating individual structures~\cite{zhao2025dymorph}.

Classical post-processing approaches attempt to bridge this gap by splitting semantic masks using heuristic operations such as watershed transforms~\cite{rambabu2007efficient}, skeletonization~\cite{zhang1984fast}, or morphological filtering~\cite{haralick1987image}. While these methods do not require instance labels, they rely on fixed rules and are highly sensitive to shape variability, leading to under-segmentation or over-segmentation in complex kidney structures (Fig.~\ref{fig:Problem}a). In contrast, learning-based methods trained with instance annotations can achieve accurate separation~\cite{bouteldja2021deep,jiang2021deep,yang2025unsupervised}, but they depend on fully or partially labeled instance masks, which are expensive and difficult to obtain at scale (Fig.~\ref{fig:Problem}b). Thus, existing approaches face a trade-off between supervision cost and segmentation quality. More recently, learning-based semantic-to-instance methods attempt to reduce this supervision burden by leveraging discriminative feature embeddings~\cite{chen2019instance,li2023sim} or boundary-aware modeling~\cite{chen2017dcan,abera2026style,he2026instabound} under semantic supervision. Although these approaches improve instance separability, they are often developed for natural images or relatively homogeneous cellular objects and lack explicit morphology-aware geometric modeling, limiting their generalization to the heterogeneous and structurally complex kidney units encountered in pathology images~\cite{deng2024prpseg}.

In this work, we incorporate \emph{morphology priors of kidney structures} into the learning framework to enable \emph{instance-mask-free instance segmentation under semantic supervision}. We propose \textbf{MORI-Seg}, which decomposes connected semantic regions into distinct instances without requiring any instance-level annotations. Instead of relying on heuristic splitting or weak instance supervision, MORI-Seg explicitly learns complementary geometric representations derived from medical object morphology, including object-centric distance fields that encode interior structure and boundary-band representations that capture contact interfaces. A class-conditioned feature disentanglement module further promotes intra-instance coherence and inter-instance separation. By jointly modeling feature-level and morphology-aware geometric cues in an end-to-end architecture, MORI-Seg achieves reliable instance separation from semantic masks alone (Fig.~\ref{fig:Problem}c).

Our contributions are threefold:
\begin{itemize}
\item We enable instance segmentation from semantic supervision alone, achieving instance-mask-free instance segmentation without requiring any instance-level annotations.
\item We incorporate explicit morphology priors of kidney structures into the learning framework, modeling complementary object-centric distance fields and boundary-band representations to guide instance separation.
\item We demonstrate through extensive cross-dataset evaluation that morphology-aware geometric learning consistently improves instance separation accuracy and morphometric reliability compared with heuristic and existing semantic-to-instance approaches.
\end{itemize}

\section{Methods}

\begin{figure*}[t]
    \centering
    \includegraphics[width=0.9\linewidth]{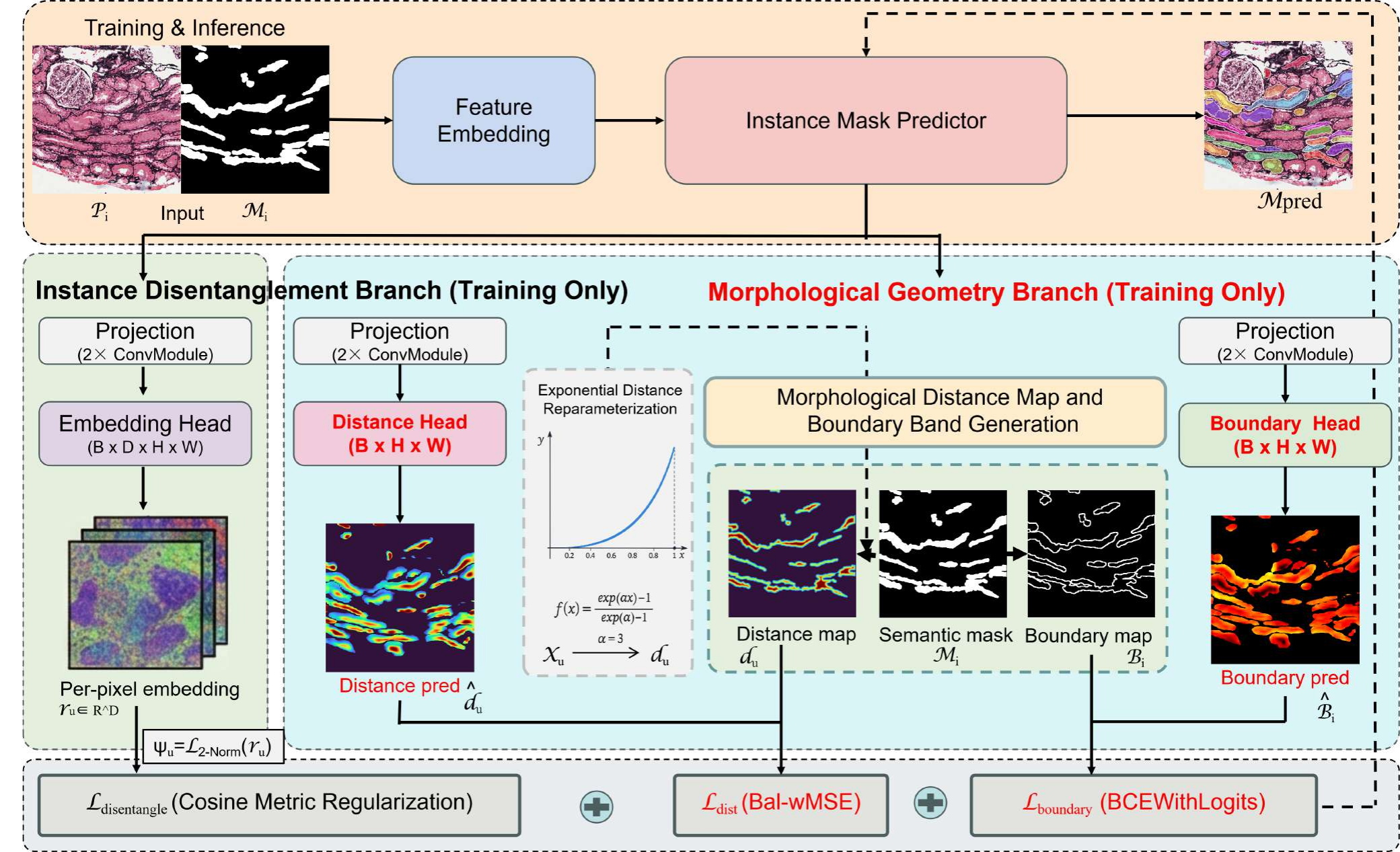}
    \caption{\textbf{Overview of MORI-Seg.} 
Built upon RTMDet-Ins, MORI-Seg introduces a Morphological Geometry Branch during training to encode kidney structure priors under semantic supervision. The branch jointly models an object-centric exponential distance field to capture interior geometry and a morphology-derived boundary-band map to represent contact interfaces between adjacent structures. These complementary geometric signals guide robust instance separation from semantic masks alone. All auxiliary heads are removed at inference, preserving the original RTMDet-Ins decoding pipeline without additional computational overhead.}
\label{fig:Method}
\end{figure*}

As illustrated in Fig.~\ref{fig:Method}, MORI-Seg learns instance segmentation from semantic supervision by jointly modeling feature-level disentanglement and morphological geometry. Our framework consists of three components: (1) instance-level feature disentanglement, inspired by pixel-embedding based instance grouping and discriminative learning paradigms; (2) morphology-aware geometric modeling via distance and boundary priors; and (3) joint optimization under semantic-only supervision. All auxiliary branches are used only during training and removed at inference, preserving the native RTMDet-Ins pipeline~\cite{lyu2022rtmdet}.

\subsection{Feature-Level Instance Disentanglement}

Kidney pathology images often contain densely packed and adherent structures. When trained solely with semantic masks, conventional segmentation models tend to merge adjacent instances. To encourage instance-level separability, we introduce an Instance Disentanglement Branch (Fig.~\ref{fig:Method}, left), which operates on the shared mask features $\mathcal{F}_{mask}$, inspired by~\cite{chen2019instance}.

This branch uses a compact projection module composed of two stacked $3\times3$ convolutional layers. At each pixel $u$, it outputs a $D$-dimensional vector $\mathbf{r}_u$, which is then $\ell_2$-normalized to obtain the embedding $\boldsymbol{\psi}_u\in\mathbb{R}^{D}$. For instance $m$ with pixel set $\mathcal{S}_m$, we further compute a normalized instance prototype $\boldsymbol{\pi}_m$:
\begin{equation}
\boldsymbol{\psi}_u=\frac{\mathbf{r}_u}{\lVert \mathbf{r}_u\rVert_2},
\qquad
\boldsymbol{\pi}_m=
\frac{\sum_{u\in\mathcal{S}_m}\boldsymbol{\psi}_u}
{\left\lVert \sum_{u\in\mathcal{S}_m}\boldsymbol{\psi}_u\right\rVert_2}.
\end{equation}

To promote intra-instance compactness, we adopt a cosine-based consistency term that pulls each pixel embedding toward its corresponding instance prototype. Since the primary failure case arises when structures touch or lie in close proximity, we restrict the separation constraint to locally neighboring instance pairs. Specifically, we construct the neighbor set $\mathcal{A}$ using a locality radius derived from neighbor\_distance and encourage prototypes associated with pairs in $\mathcal{A}$ to be approximately orthogonal. The resulting disentanglement objective can be written in expanded form as:
\begin{equation}
\mathcal{L}_{\mathrm{disentangle}}
=
\frac{1}{M}\sum_{m=1}^{M}\frac{1}{|\mathcal{S}_m|}\sum_{u\in\mathcal{S}_m}
\left(1-\cos\!\left(\boldsymbol{\psi}_u,\boldsymbol{\pi}_m\right)\right)
\;+\;
\lambda_{\mathrm{sep}}\cdot
\frac{1}{|\mathcal{A}|}\sum_{(m,n)\in\mathcal{A}}
\left|\boldsymbol{\pi}_m^\top\boldsymbol{\pi}_n\right|.
\end{equation}
where $M$ is the number of instances involved in the computation, $\mathcal{S}_m$ denotes the pixel set of instance $m$ with $|\mathcal{S}_m|$ pixels, $\mathcal{A}$ is the set of neighboring instance pairs with $|\mathcal{A}|$ pairs, $\lambda_{\mathrm{sep}}$ is the weight of the separation term, and $D$ is the embedding dimension.

\subsection{Morphological Geometry Branch}

While feature separation provides identity-level cues, reliable instance separation also requires geometric priors. We therefore introduce a Morphological Geometry Branch (Fig.~\ref{fig:Method}, middle and right), which learns complementary interior and interface representations through two geometric signals: an object-centric distance field and a boundary-band map.

\noindent\textbf{Distance Map Prior.}

The Distance Head produces a scalar distance prediction $\hat d_u$ at each location $u$ on the feature grid $\Omega$. Supervision is obtained directly from the semantic mask by treating each connected foreground region as an instance region, denoted by $\{\mathcal{M}_i\}_{i=1}^{M}$. For any pixel $u\in\mathcal{M}_i$, we define a normalized Euclidean distance-to-boundary target $x_u\in[0,1]$ as
\begin{equation}
x_u=
\frac{\min\limits_{v\in\partial\mathcal{M}_i}\|u-v\|_2}
{\max\limits_{u'\in\mathcal{M}_i}\min\limits_{v\in\partial\mathcal{M}_i}\|u'-v\|_2+\varepsilon},
\qquad u\in\mathcal{M}_i,
\end{equation}
The index $u'$ spans all pixels within $\mathcal{M}_i$ to determine the normalization factor. Consequently, $x_u$ approaches $0$ near the boundary and increases progressively toward the region interior.

In crowded fields, regression can be disproportionately influenced by boundary-adjacent noise and minor label ambiguity, which compresses the effective contrast between cores and borders. To obtain a more stable geometric signal, we reshape the target with an exponential reparameterization:
\begin{equation}
d_u=f(x_u)=\frac{\exp(\alpha x_u)-1}{\exp(\alpha)-1},
\qquad \alpha=3.
\end{equation}
This transformation keeps the target in $[0,1]$ while redistributing sensitivity across the region: variations near the boundary are damped, whereas interior values preserve a clearer dynamic range. As a result, the Distance Head is encouraged to learn a cleaner monotonic decay from centre to boundary, providing a more discriminative cue for separating touching instances.

Distance regression is optimized using a \emph{foreground--background balanced weighted mean squared error} (Bal-wMSE):
\begin{equation}
\mathcal{L}_{\mathrm{dist}}=\mathrm{Bal\text{-}wMSE}(\hat d_u,d_u),
\end{equation}
where Bal-wMSE refers to a pixel-wise MSE with balanced weighting between foreground and background, implemented via cardinality normalization, to prevent either set from dominating the objective.

\noindent\textbf{Boundary Map Prior.}
Distance cues alone may be insufficient when instances share long, low-contrast interfaces. In kidney pathology, structures exhibit substantial variability in size and topology, with irregular and elongated morphologies rather than compact, cell-like shapes. Under such conditions, purely object-centric distance maps may not adequately model complex contact regions. 

We therefore construct an explicit boundary-band supervision signal. For each semantic mask $\mathcal{M}_i$, a thick boundary map is generated using a morphological gradient:
\begin{equation}
\mathcal{B}_i = \mathbf{1}\left[\mathrm{Dilate}(\mathcal{M}_i,w)-\mathrm{Erode}(\mathcal{M}_i,w)>0\right],
\end{equation}
and aggregated across instances:
\begin{equation}
\mathcal{B}=\bigvee_i \mathcal{B}_i.
\end{equation}
A Boundary Head predicts $\hat{\mathcal{B}}$, supervised with a reweighted binary cross-entropy loss:
\begin{equation}
\mathcal{L}_{boundary}=\mathrm{BCEWithLogits}(\hat{\mathcal{B}},\mathcal{B}).
\end{equation}
This branch explicitly models contact interfaces, complementing the interior distance prior.

\subsection{Joint Geometry-Feature Learning under Semantic Supervision}

The final objective integrates feature disentanglement and geometric modeling:
\begin{equation}
\mathcal{L}_{total}
=\mathcal{L}_{rtmdet}
+\lambda_{feaure}\mathcal{L}_{disentangle}
+\lambda_{reg}\mathcal{L}_{dist}
+\lambda_{bd}\mathcal{L}_{boundary}.
\label{eq:ltotal}
\end{equation}

By jointly learning instance-discriminative embeddings and morphology-aware geometric representations from semantic masks alone, MORI-Seg decomposes connected semantic regions into distinct instances without requiring any instance-level annotations. Importantly, all auxiliary heads are removed during inference, ensuring that the final model retains the original RTMDet-Ins decoding pathway without additional computational overhead.

\section{Data and Experiments}

\textbf{Data.} Experiments are conducted on two complementary kidney pathology datasets. 
\textbf{KI} (training) provides semantic-level annotations of arteries (ART), peritubular capillaries (PTC), distal tubules (DT), proximal tubules (PT), and glomerular capsules (CAP), curated from the NEPTUNE project~\cite{NEPTUNE}. 
\textbf{KPMP} (evaluation) contains 385 PAS-stained whole-slide images with predefined instance-level annotations for arteries\_arterioles, peritubular-capillaries, tubules, and non-globally-sclerotic glomeruli~\cite{KPMP}. 
A subset of 66 slides with complete instance annotations for the four target categories is used exclusively for evaluation; no KPMP data are used during training. 
To ensure consistent evaluation, categories are aligned as follows: ART $\rightarrow$ arteries(arterioles), PTC $\rightarrow$ peritubular capillaries, DT+PT $\rightarrow$ tubules, and CAP $\rightarrow$ non globally sclerotic glomeruli.

\textbf{Experimental Setup.} MORI-Seg is trained on KI using semantic supervision only with an 8:2 train/validation split. Fixed-size \(512\times512\) patches are extracted using a class-specific magnification strategy: PTC at \(40\times\), and ART, CAP, DT, and PT at \(10\times\). Evaluation on KPMP follows the same magnification protocol. Performance is measured using COCO-style instance segmentation mAP, with non-maximum suppression applied at inference. All experiments are conducted on a workstation equipped with an NVIDIA RTX A6000 GPU.

\begin{table*}
\caption{Instance segmentation performance. Each entry is reported as mAP / mAP50 / mAP75.}
\centering
\footnotesize
\setlength{\tabcolsep}{4pt}
\renewcommand{\arraystretch}{1.1}
\begin{adjustbox}{max width=\textwidth}
\begin{tabular}{l c c c c c}
\toprule
Method & ART & CAP & PTC & DT+PT & Average \\
\midrule

Watershed~\cite{rambabu2007efficient}
& 0.012/0.026/0.011
& 0.172/0.278/0.175
& 0.015/0.046/0.009
& 0.078/0.167/0.071
& 0.070/0.129/0.067 \\

Skeleton~\cite{zhang1984fast}
& 0.012/0.028/0.010
& 0.092/0.178/0.083
& 0.019/0.050/0.012
& 0.066/0.141/0.056
& 0.047/0.099/0.040 \\

Morphology~\cite{haralick1987image}
& 0.111/0.190/0.114
& 0.449/0.596/0.497
& 0.083/0.179/0.066
& 0.235/0.376/0.246
& 0.219/0.335/0.231 \\

\midrule

CBNet-Swin-Tiny~\cite{liang2022cbnet}
& 0.121/0.198/0.129
& 0.323/0.406/0.372
& 0.124/0.257/0.109
& 0.325/0.467/0.369
& 0.223/0.332/0.245 \\

CondInst-R50~\cite{tian2020conditional}
& 0.101/0.165/0.110
& 0.439/0.652/0.516
& 0.123/0.248/0.112
& 0.275/0.406/0.310
& 0.235/0.368/0.262 \\

SOLOv2-EfficientNet~\cite{wang2020solov2}
& 0.025/0.040/0.026
& 0.329/0.468/0.384
& 0.111/0.226/0.100
& 0.227/0.413/0.228
& 0.173/0.287/0.185 \\

SOLOv2-MobileNetV2~\cite{wang2020solov2}
& 0.016/0.024/0.015
& 0.315/0.487/0.356
& 0.099/0.208/0.086
& 0.103/0.227/0.080
& 0.133/0.237/0.134 \\

YOLOv12~\cite{tian2025yolov12}
& 0.039/0.069/0.041
& 0.431/0.564/0.511
& 0.013/0.028/0.012
& 0.130/0.249/0.127
& 0.153/0.227/0.173 \\

DINOv3-Mask2Former~\cite{simeoni2025dinov3}
& 0.072/0.107/0.078
& 0.274/0.341/0.312
& 0.120/0.245/0.106
& 0.267/0.402/0.297
& 0.183/0.274/0.198 \\

RTMDet-Ins~\cite{lyu2022rtmdet}
& 0.162/0.253/0.180
& 0.490/0.621/0.551
& 0.113/0.233/0.101
& 0.501/0.737/0.565
& 0.317/0.461/0.350 \\

\midrule

SCD~\cite{wang2025distilling}
& 0.018/0.045/0.010
& 0.003/0.006/0.001
& 0.034/0.098/0.012
& 0.098/0.263/0.022
& 0.038/0.103/0.011 \\

InstaBound~\cite{he2026instabound}
& 0.029/0.045/0.031
& 0.000/0.000/0.000
& 0.061/0.154/0.041
& 0.000/0.000/0.000
& 0.023/0.050/0.018 \\

DCAN~\cite{chen2017dcan}
& 0.073/0.199/0.040
& 0.496/0.661/0.564
& 0.036/0.125/0.013
& 0.175/0.561/0.041
& 0.195/0.386/0.164 \\

Pixel-Embedding~\cite{chen2019instance}
& 0.155/0.228/0.172
& 0.329/0.433/0.379
& 0.133/0.265/0.122
& 0.509/0.723/0.578
& 0.282/0.412/0.313 \\

Style-Boundary~\cite{abera2026style}
& \textbf{0.214}/\textbf{0.321}/0.240
& 0.524/0.699/0.600
& 0.144/0.286/0.132
& 0.513/0.734/0.585
& 0.349/0.510/0.390 \\

SIM~\cite{li2023sim}
& 0.214/0.315/\textbf{0.244}
& 0.533/0.710/0.609
& 0.147/0.289/0.136
& 0.523/0.746/0.599
& 0.354/0.515/0.397 \\

MORI-Seg (Ours)
& 0.198/0.286/0.230
& \textbf{0.611}/\textbf{0.799}/\textbf{0.712}
& \textbf{0.152}/\textbf{0.294}/\textbf{0.145}
& \textbf{0.596}/\textbf{0.820}/\textbf{0.691}
& \textbf{0.389}/\textbf{0.550}/\textbf{0.444} \\

\bottomrule
\end{tabular}
\end{adjustbox}
\label{tab:instseg_map_compact}
\end{table*}

\begin{figure*}
    \centering
    \includegraphics[width=1.0\linewidth]{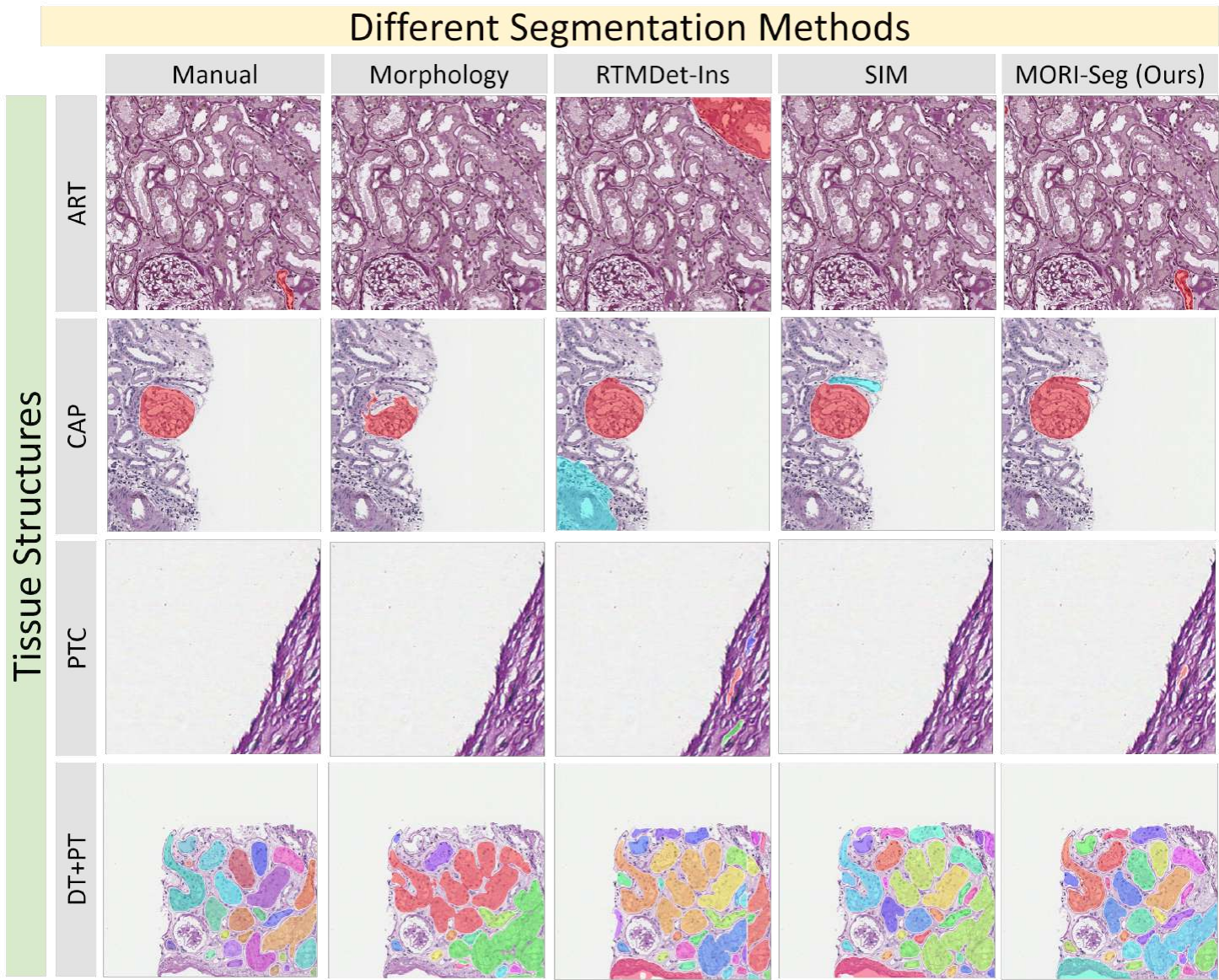}
    \caption{Qualitative comparison of instance segmentation results across four kidney tissue structures, showing that MORI-Seg achieves more accurate instance separation and boundary delineation in crowded and morphologically complex regions compared with heuristic post-processing and representative learning-based methods.}
\label{fig:result}
\end{figure*}

\section{Results}

We compare MORI-Seg with three groups of baselines: (1) heuristic post-processing methods applied to semantic segmentation (Omni-Seg~\cite{deng2023omni} + watershed~\cite{rambabu2007efficient}, skeleton~\cite{zhang1984fast}, morphology~\cite{haralick1987image}), (2) fully supervised instance segmentation backbones (CBNet-Swin-Tiny~\cite{liang2022cbnet}, CondInst-R50~\cite{tian2020conditional}, SOLOv2 variants~\cite{wang2020solov2}, YOLOv12~\cite{tian2025yolov12}, DINOv3-Mask2Former~\cite{simeoni2025dinov3}, RTMDet-Ins~\cite{lyu2022rtmdet}), and (3) semantic-to-instance learning approaches (SCD~\cite{wang2025distilling}, InstaBound~\cite{he2026instabound}, DCAN~\cite{chen2017dcan}, Pixel-Embedding~\cite{chen2019instance}, Style-Boundary~\cite{abera2026style}, SIM~\cite{li2023sim}).

Table~\ref{tab:instseg_map_compact} and Fig.~\ref{fig:result} present quantitative and qualitative comparisons on the KPMP test subset across four kidney structure categories (ART, CAP, PTC, and DT+PT). Heuristic post-processing methods show limited separation capability and are prone to under- or over-segmentation in crowded regions, while fully supervised instance segmentation backbones improve performance but require instance-level annotations. Under semantic-only supervision, MORI-Seg achieves the strongest overall performance and demonstrates consistent improvements across categories, particularly for morphologically complex and densely packed structures such as CAP, DT+PT, and PTC. Qualitative results further confirm that MORI-Seg produces cleaner instance boundaries and more reliable separation in adherent regions, highlighting the benefit of morphology-aware geometric priors for robust instance decomposition without instance annotations.

\noindent\textbf{Ablation Study}
Table~\ref{tab:ablation_rtm_ablation} shows that adding a distance prior consistently improves performance over the baseline, validating the effectiveness of object-centric geometric supervision, with the exponential mapping providing a slight but consistent advantage over the quadratic form. Incorporating boundary-band supervision further enhances results, demonstrating the complementary role of interface modeling. The full MORI-Seg configuration achieves the strongest performance, indicating that jointly modeling interior geometry and contact boundaries leads to more robust instance separation under semantic-only supervision.

\begin{table}[t]
\caption{Ablation study of MORI-Seg components on RTMDet-Ins.}
\centering
\footnotesize
\setlength{\tabcolsep}{4pt}
\renewcommand{\arraystretch}{1.08}
\resizebox{\linewidth}{!}{
\begin{tabular}{lccc|cc|ccc}
\toprule
\multirow{2}{*}{Variant} 
& \multicolumn{2}{c}{Distance Prior} 
& \multirow{2}{*}{Boundary Prior} 
& \multirow{2}{*}{$\Delta$Params} 
& \multirow{2}{*}{$\Delta$Time} 
& \multicolumn{3}{c}{Average} \\
\cmidrule(lr){2-3} \cmidrule(lr){7-9}
& \textsc{Pow2} 
& \textsc{Exp3} 
& 
& 
& 
& mAP 
& mAP50 
& mAP75 \\
\midrule

Baseline (RTMDet-Ins)
&  &  &  
& 57.309M 
& 10:02:08
& 0.317 & 0.461 & 0.350 \\

+ Distance Prior (\textsc{Pow2})
& \checkmark &  &  
& +0.012M
& +02:28:40
& 0.370 & 0.530 & 0.423 \\

+ Distance Prior (\textsc{Exp3})
&  & \checkmark &  
& +0.012M
& +02:34:12
& 0.372 & 0.534 & 0.423 \\

+ Boundary Prior
&  &  & \checkmark 
& +0.012M
& +02:46:30
& 0.376 & 0.537 & 0.430 \\

MORI-Seg (Ours)
&  & \checkmark & \checkmark 
& +0.024M
& +02:51:24
& \textbf{0.389} & \textbf{0.550} & \textbf{0.444} \\

\bottomrule
\end{tabular}
}
\label{tab:ablation_rtm_ablation}
\end{table}
\section{Conclusion}

We presented MORI-Seg, an instance-mask-free framework that learns instance segmentation from semantic supervision by modeling morphological geometry. By jointly integrating instance-discriminative feature embeddings with complementary geometric priors, including object-centric distance fields and boundary-band representations, MORI-Seg decomposes connected semantic regions into structurally consistent instances without requiring any instance-level annotations. Extensive experiments demonstrate consistent improvements over classical post-processing and existing semantic-to-instance approaches. By eliminating the trade-off between supervision cost and segmentation quality, MORI-Seg provides a scalable solution for transforming widely available semantic annotations into reliable instance-level representations, and offers a generalizable geometry-driven paradigm for medical image analysis tasks where instance annotations are scarce.

\noindent\textbf{Acknowledgements}. 
This research is supported by the WCM Radiology AIMI Fellowship and the WCM CTSC Pilot Award. This research is also supported by NIH R01DK135597 (Huo), NSF 2434229 (Huo), and DoD HT9425-23-1-0003 (HCY)

\noindent\textbf{Disclosure of Interests}. The authors have no competing interests.
%
%
 \bibliographystyle{splncs04}
 \bibliography{main}
%




\end{document}